\documentclass[journal,onecolumn]{IEEEtran}
\IEEEoverridecommandlockouts
\usepackage{cite}
\usepackage{amsmath,amssymb,amsfonts}
\usepackage{algorithmic}
\usepackage{graphicx,float}
\usepackage{textcomp}
\usepackage{xcolor}
\usepackage{csquotes}
\usepackage{subcaption}
\usepackage{multicol}
\usepackage{lipsum}
\usepackage{tabularx,multirow,booktabs,blindtext}
\def\BibTeX{{\rm B\kern-.05em{\sc i\kern-.025em b}\kern-.08em
    T\kern-.1667em\lower.7ex\hbox{E}\kern-.125emX}}

\usepackage{authblk}

\newcommand*{\affaddr}[1]{#1} 
\newcommand*{\affmark}[1][*]{\textsuperscript{#1}}
\newcommand*{\email}[1]{\textit{#1}}

\begin{document}

\title{Fast Fourier Transformation for Optimizing Convolutional Neural Networks in Object Recognition 
}

\author{%
Varsha Nair\affmark[1], Moitrayee Chatterjee\affmark[1], Neda Tavakoli\affmark[2], Akbar Siami Namin\affmark[1], and Craig Snoeyink\affmark[3]\\
\affaddr{\affmark[1, 2]Department of Computer Science,}
\affaddr{\affmark[3]Department of Mechanical and Aerospace Engineering}\\
\affaddr{\affmark[1]Texas Tech University,}
\affaddr{\affmark[2]Georgia Institute of Technology,}
\affaddr{\affmark[3]State University of New York at Buffalo} \\
\email{\{varsha.nair, moitrayee.chatterjee, akbar.namin\}@ttu.edu; neda.tavakoli@gatech.edu; craigsno@buffalo.edu}\\
}

\maketitle

\begin{abstract}

This paper proposes to use Fast Fourier Transformation-based U-Net (a refined \enquote{fully convolutional networks}) and perform image convolution in neural networks. Leveraging the Fast Fourier Transformation, it reduces the image convolution costs involved in the Convolutional Neural Networks (CNNs) and thus reduces the overall computational costs.
The proposed model identifies the object information from the images. We apply the Fast Fourier transform algorithm on an image data set to obtain more accessible information about the image data, before segmenting them through the 
U-Net architecture. More specifically, we implement the FFT-based convolutional  neural  network to improve the training time of the network. The proposed approach was applied to publicly available Broad Bioimage Benchmark Collection (BBBC) dataset. Our model demonstrated  improvement in training time during convolution from $600-700$ ms/step to $400-500$ ms/step. We evaluated the accuracy of our model using Intersection over Union (IoU) metric showing
significant improvements\footnote{This paper is the pre-print of a paper to appear in the proceedings of the IEEE International Conference on Machine Learning Applications (ICMLA'20) entitled: ``{\it Optimizing CNN using Fast Fourier Transformation for Object Recognition}.''}.

\end{abstract}
\begin{IEEEkeywords}
Object Recognition, CNN, Fast Fourier Transformation, Image Processing.
\end{IEEEkeywords}

\section{Introduction}

Computer vision follows a sequence of operations to identify the characteristics of an image to perceive and analyze. Computer vision can be extensively used to solve a wide variety of problems in the field of remote sensing, spectroscopy, medical imaging, image processing, meteorology, astronomy and microscopy. Image recognition is an exciting research fields in computer vision, for which researchers predict that the global market will reach $\$38.92$ billion by 2021 \cite{ImageRec74:online}.

The application of deep learning approaches for image enhancement, restoration and morphing \cite{he2016deep, krizhevsky2012imagenet}.
has significantly improved the data (or features) extraction from images. Moreover, Convolutional Neural Networks (CNNs) have been extremely successful in image processing. However, CNNs are computationally expensive due to computation costs of convolutions. By transforming images and CNN kernels into frequency space using Fast Fourier Transformation (FFT), a convolution operator becomes a single element-wise multiplication. Hence, FFT is used tremendously for image convolution. But, there is an overhead cost involved in performing FFT. This type of transformation is efficient in reducing the convolution cost only when the convolution is large. Recently, convolutions of CNNs have been optimized using FFT~\cite{vasilyev2015cnn} and being adapted in mobile and embedding systems. 
Here, we apply FFT to the input data only and demonstrate the performance improvement on detecting objects. 

Convolutional Networks \cite{lecun1989backpropagation} have existed for some times but the required size of training sets and the network size restricted their wide adoptions. On the other hand, their deep variations have shown outstanding performance in visual object recognition tasks \cite{krizhevsky2012imagenet}, through supervised training. 

The object identification algorithms typically use extracted features and learning algorithms to apprehend instances of an object category. Currently, most of the object detection models specify where object is located in image with a bounding box \cite{redmon2017yolo9000,he2017mask}. 

This study attempts to identify object position using a FFT-based convolutional neural network to detect the position by image segmentation rather than a bounding box. One of the major issues with using CNN for image dataset is the training time. This study focuses on lesion detection in image data set using the U-Net model and thus aims at improving the training time by implementing Fast Fourier Transformation layer to the U-Net model. 
This paper proposes to accomplish this by using a Fast Fourier convolution with fully CNN. The proposed model detects objects from an image, utilizing the scaling properties of FFT that improves the convolution time. The key contributions of this paper are as follows:
\begin{itemize}
    \item Introducing Fast Fourier Transformation-based deep learning algorithm (i.e., FFT-based CNN) in the context of object recognition for optimization purposes,
    \item Implementing FFT on an U-Net architecture and capturing key features, and
    \item Evaluating the proposed Fast Fourier Transformation-based CNN through a set of experiments using bio-image dataset in the domain of health analytic.
\end{itemize}

The rest of the paper is organized as follows: we review the existing work in object identification in Section \ref{sec:rel} and then provide explanation of the ideas and theories used in our implementation in Section \ref{sec:bck}. Section \ref{sec:meth} presents the methodology and generic algorithm for proposed approach. Section \ref{sec:cs} gives an insight to the experiments and the results. We concluded the paper in Section \ref{sec:con}.

\section{Related Work}
\label{sec:rel}


To incorporate the contextual information for lesion recognition Gao et al. \cite{gao2015automatic} proposed a fusion of CNN (convolutional Neural Networks) and RNN (Recurrent Neural Networks) based approach. Shen et al.  \cite{shen2015automatic} implemented three CNNs for addressing the problem of object recognition. While these approaches are targeted for object identification for 2D images, Setio et al. \cite{setio2016pulmonary} proposed a multi-stream CNN for object classification in 3D images.

For quantitative analysis of clinical characteristics, the bio-image needs segmentation to identify the pixels that represent the object of interest. Ronneberger et. al.\cite{RFB15a} proposed U-Net for bio medical image segmentation. It is a refined architecture developed from fully-convolutional neural network which is used for fast and precise segmentation of images.  It was named on the fact that its architecture is in the shape of the letter \enquote{U}. The U-Net has been further extended \cite{cciccek20163d} for 3D image segmentation. V-Net \cite{milletari2016v} is another variant of U-Net for 3D images. RNNs are also used for image segmentations \cite{xie2016spatial}. Stollenga et al. \cite{stollenga2015parallel} implemented a 3D LSTM-RNN with convolutional layer. Andermatt et al. \cite{andermatt2016multi} proposed gated RNN for segmenting 3D bio-images. Chen et al. \cite{chen2016combining} combined RNN with U-Net like architecture for segmentation. Convolutional Neural Netowrks also have been used in many other application domains such as mapping genome data into images \cite{tavakoli2020seq2image}.

\section{Theoretical Background}
\label{sec:bck}

This section discusses the two major techniques that we used for the convolution and image processing: (1) Fast Fourier Transformation and (2) U-Net.
\subsection{Fast Fourier Transformation}
In image processing, FFT is applied on an image to decompose it into real and imaginary components, representing the image in a frequency domain\cite{goodmanIntroductionFourierOptics2017}. 
The FFT of a 2D image \cite{muthyalampalli2009implementation} can be calculated using Equations (\ref{eq:fft}) and (\ref{eq:ifft}):

\begin{equation}
\label{eq:fft}
    F(x,y) = \sum \limits_{m=0}^{M-1} \sum \limits_{n=0}^{N-1} f(m,n) e^{-(i \times 2 \times \pi (x\frac{m}{M}+y \frac{n}{N}))}
\end{equation}


\begin{equation}
\label{eq:ifft}
    f(x,y) = \frac{1}{M.N}\sum \limits_{m=0}^{M-1} \sum \limits_{n=0}^{N-1} F(m,n) e^{(i \times 2 \times \pi (x\frac{m}{M}+y \frac{n}{N}))}
\end{equation}

where $f (m, n)$ represents the pixel at position $(m, n)$, whereas $F(x, y)$ is the function to represent the image in the frequency domain pertaining to the position $x$ and $y$, $M \times N$ represents dimension of the image, and $i$ is $sqrt(-1)$ \cite{goodmanIntroductionFourierOptics2017}. 

In our work, we used a tensorflow-based implementation of Fast Fourier Transform in order to produce a transformed feature map in Fourier domain, which is then provided to U-Net for object segmentation.
\subsection{U-Net Architecture}
Ronneberger et al. proposed the U-Net architecture \cite{RFB15a} consists of a contracting path to apprehend the context and a symmetric expanding path that enables precise localization. This network is trained end-to-end using very few images and outperformed  sliding-window convolutional network like OverFeat \cite{sermanet2013overfeat}.

\begin{figure}[h]
\includegraphics[width=\linewidth]{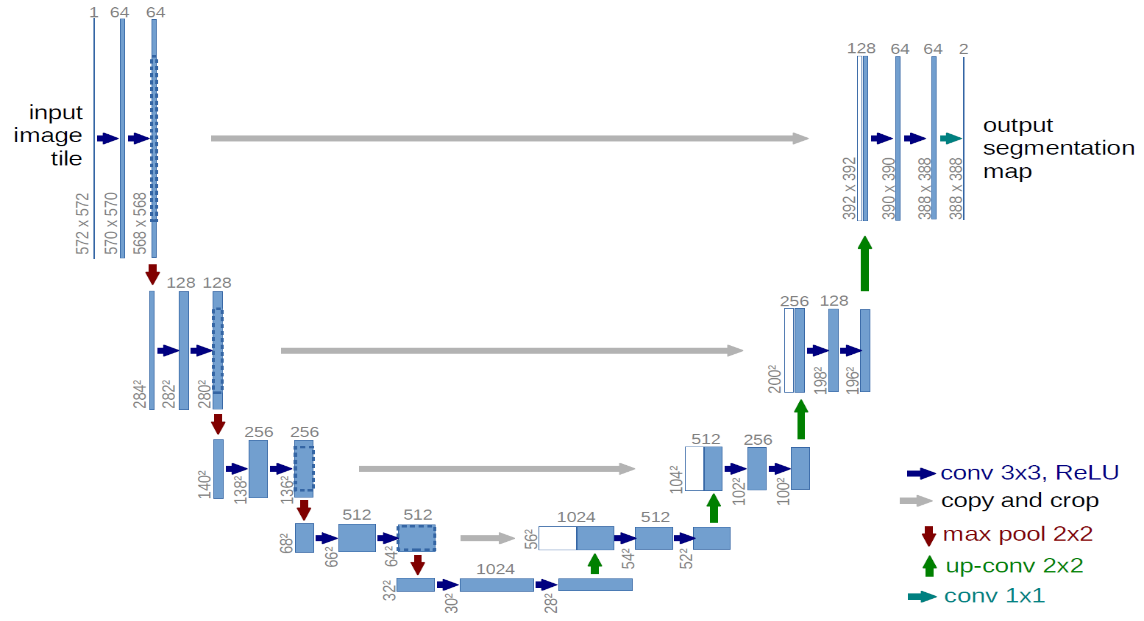}
\caption{Architecture of U-Net \cite{RFB15a} on lowest resolution of $32\times32$ pixels.}
\label{fig:UNetarch}
\end{figure}

In Figure \ref{fig:UNetarch}, the architecture of U-net is presented where each blue box is a multi-channel feature map. The number of channels is indicated on the top of each box. White boxes are mapping of copied features. The arrows denote the various operations within the network.  

\section{Methodology}
\label{sec:meth}

This section presents the architecture of the proposed model along with the developed algorithm. This work implements the U-Net \cite{RFB15a} with Fast Fourier Transform-based NN. 
The neural network implements the Fast Fourier Transform for the convolution between image and the kernel (i.e., mask). We trained the NN with labeled dataset \cite{ljosa2012annotated} which consists of synthetic cell images and masks. Each pixel on the image is classified as either being part of a cell or not. The image is a matrix that has values ranging between $0$ and $255$ (RGB). The mask (i.e., kernel) is a 2-D matrix with white and black values. Each pixel in the predicted mask represents the probability that the pixel is part of a cell. The output of the prediction model is clustered to identify the object location in a 2-D environment.

\subsection{Model Overview}
The proposed architecture is built upon the U-Net, which is a fully Convolutional Network. The U-Net is modified to yield better segmentation in medical imaging. The proposed model consists of the contracting path and an expansion path. These two together form a U-shaped architecture. The contracting path comprises of a fully connected convolutional network and includes repeated application of convolutions, each of which is followed by an exponential Linear Unit (eLU) and max-pooling operation. The contraction path decreases the spatial information, while increasing the feature information. The expansive path combines the feature and spatial information through a sequence of up-convolutions and concatenations with high-resolution features from the contracting path. The up-convolution in this experiment is de-convolution operation, which is used for up-sampling. In this model, we incorporate dropout in every block to avoid overfitting.  Figure \ref{fig:archModel} displays the architecture of the proposed model. \\
It consists of:

\begin{figure}[!t]
\includegraphics[width=\linewidth]{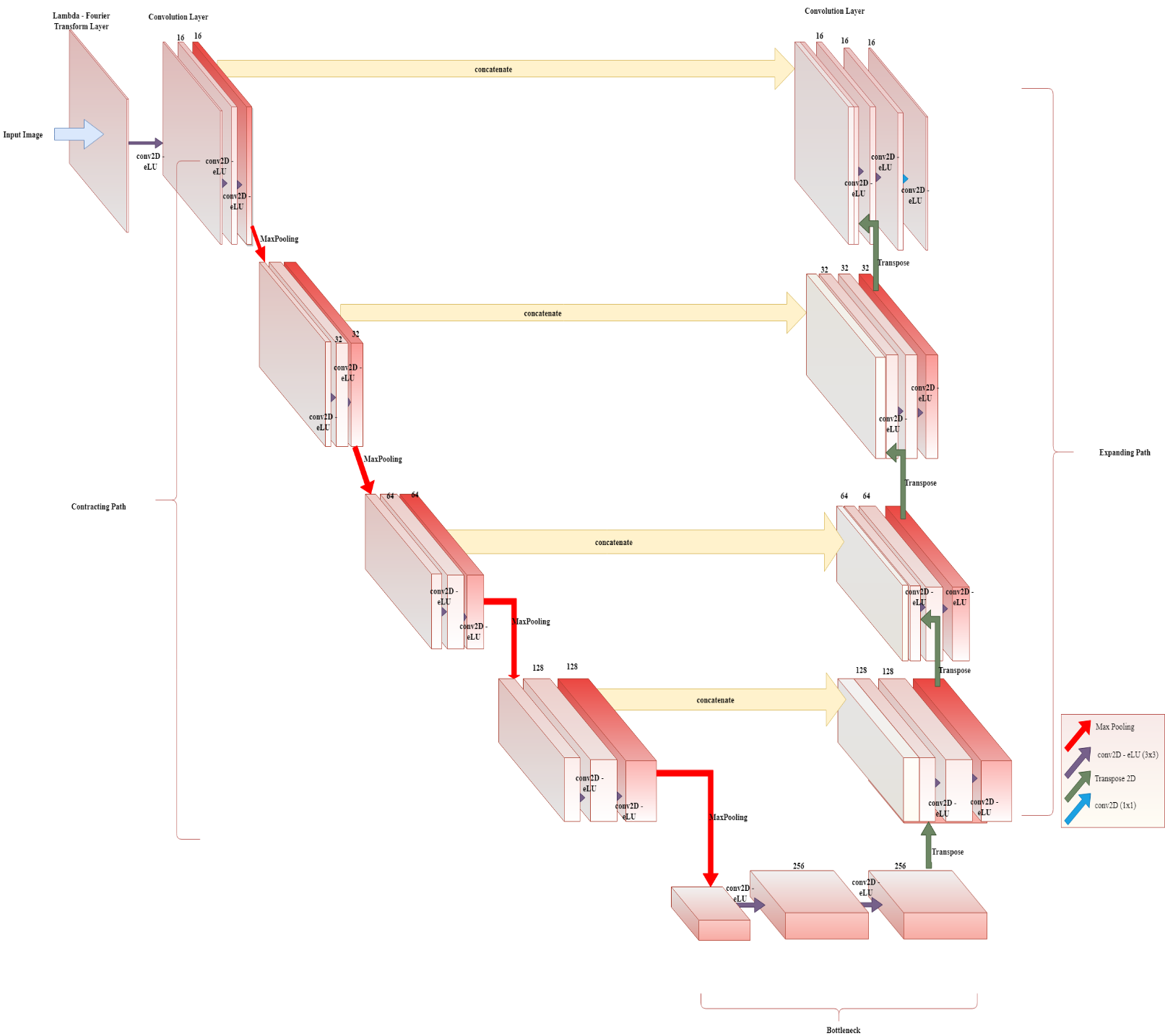}
\caption{Architecture of the proposed model.}
\label{fig:archModel}
\end{figure}

\begin{enumerate}
    \item \textbf{Input Layer.} This layer takes the input image and performs Fast Fourier convolution by applying the Keras-based FFT function\cite{chollet2015keras}. The input layer is composed of:
    \begin{enumerate}
        \item A lambda layer with Fast Fourier Transform
        \item A 3x3 Convolution layer and activation function, and
        \item A lambda layer with Inverse Fast Fourier Transform.
    \end{enumerate}

\item \textbf{Contracting Path.} The Contracting layer consists of four blocks where each block is a group of operations bound together logically. In this part of the architecture, each block entails:
    \begin{enumerate}
    \item Two $3\times3$ Convolution layers with activation function, and 
    \item A $2\times2$ Max pooling layer.
    \end{enumerate}
    
This path applies $3\times3$ convolution repeatedly. Each convolution is followed by exponential Linear Unit (eLU) and a $2\times2$ max pooling operation with stride 2 for down-sampling. Here, at every max pooling step the feature channels are doubled.

\item \textbf{Bottleneck.} The bottleneck is the connection between contracting and expanding paths. It consists of two convolutional layers with dropout:
    \begin{enumerate}
    \item Two $3\times3$ Convolution layers with activation functions, and
    \item A $2\times2$ Convolution layer.
    \end{enumerate}

\item \textbf{Expanding.} Every step in the expansive path consists of an up-sampling of the feature map, followed by a $2\times2$ convolution to reduce the number of features.  A concatenation is applied with the corresponding feature map from the contracting path. Furthermore, two $3\times3$ convolutions with eLU are applied. It consists of 4 blocks where each comprises of:
    \begin{enumerate}
\item A de-convolution layer with stride 2, 
\item A concatenation with the corresponding cropped feature map from the contracting path, 
\item Two 3x3 convolution layers with activation functions, 
\item An output layer with $1\times1$ convolutional layer, and 
\item At the final layer, a $1\times1$ convolution layer is used to map each feature vector to the input class.
\end{enumerate}
\end{enumerate}

\subsection{Process Flow}

The Figure \ref{fig:algodropout} provides a process flow of the proposed approach. 
The proposed architecture is built upon the U-Net, which is a fully Convolutional Network. The U-Net is modified to yield better segmentation in medical imaging. After importing the training dataset with image and mask for training the neural network, we resized the dimensions of the input image to $256 \times 256$ limit. Then a tensorflow spectral function is applied to the images to compute 2-D Fast Fourier Transform and Inverse Fast Fourier Transform. This tensorflow-based FFT layer is added to the neural network. The convolution of input image and mask is then implemented using a 2-D convolutional layer, which performs element-wise multiplications between the Fourier transform of the input and mask. Subsequently, an Inverse Fourier Transform layer is applied to retain the original dimensions.

The purpose of pooling in the contraction path is to reduce the size of the feature channels so that we have fewer parameters in the network. Therefore, the 
The NN retains the most relevant features (max valued pixels) from each region and discards the rest of pixels, and it is achieved through  pooling in the contraction path. The output of this layer is provided to a flattening layer, which flattens all feature structure to create a single long feature vector to be used by the output layer for object identification. 

In this model, we incorporate dropout in every block to avoid overfitting.  Figure \ref{fig:archModel} displays the architecture of the proposed model. 
\subsection{Drop Out} 
\begin{figure}[!t]
\includegraphics[width=\linewidth, height=18cm]{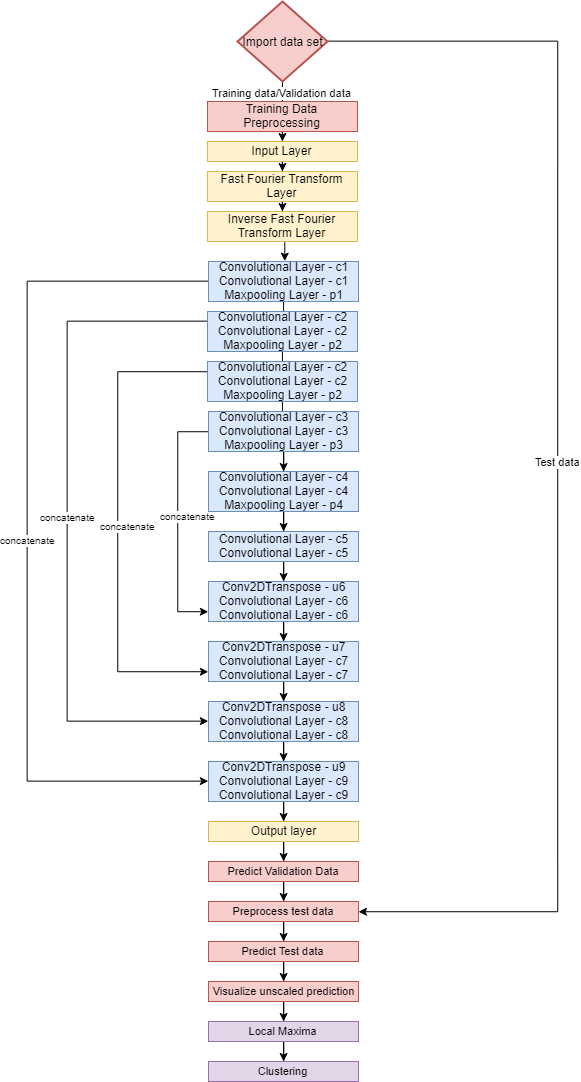}
\caption{Process flow of the proposed architecture.}
\label{fig:algodropout}
\end{figure}

Dropout is the technique of ignoring randomly selected nodes during training to prevent overfitting in neural networks. They are \enquote{dropped-out} randomly. For this experiment dropout is set on each block in the architecture. In Figure \ref{fig:algodropout}, the dropout rate is set as $0.1$ for the blocks c1, c2, c8 and c9 and $0.2$ for the blocks c3, c4, c5, c6 and c7.

\subsection{Training, Validation, and Testing}

The model is trained on the labeled data set where each image has a mask. The mask helps to identify the position of the object of interest in the main image. Additionally, validation and model fitting set is used for unbiased evaluation of the model on the training dataset. 
The model is tested against validation test data so as to ensure to calculate the model metrics. It is used to avoid overfitting. In this experiment, the validation test data consists of image and mask where the mask is compared with prediction to generate intersection over union scale.

After training the model, a scaled test set is used for an unbiased performance evaluation. The proposed model predicts the test mask from the test image dataset, which consists of the position of the objects in the image. The test mask is resized and local maxima are calculated on the test mask to capture the high intensity points. Further, DBScan clustering algorithm is applied to aggregate the local maxima in order to capture the object location and the number of objects in the image \cite{ester_density-based_1996}. 

\section{Case Study}
\label{sec:cs}

This section presents the results of a case study and implementation of the proposed FFT-based neural network algorithm. 
In this study, the accuracy of the result from the CNN based model is compared to the result of a non-CNN based system.

\subsection{Experimental Procedure and Data Collection}

Out of the total $19,200$ simulated cell microscopy images available in the  BBBC005v1\cite{ljosa2012annotated} dataset, the training set comprises of $3,600$ images and $1,200$ masks while test set comprises of $85$ images. The validation set comprises of $47$ images and $47$ masks from the training set to ensure the accuracy of the model by fitting better models. 
The model uses these processed dataset 
and predicts the mask, which is then further processed by resizing and calculating the local maxima. The processed image is clustered in order to identify the object position and number in the image.
\subsection{Results}
\begin{figure}[h]
\center
\includegraphics[width=.6\linewidth]{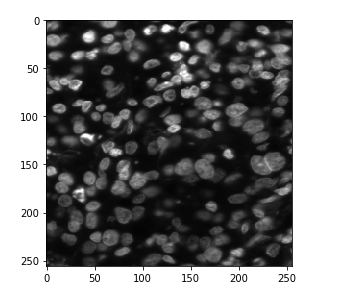}
\caption{Original image.}
\label{fig:org}
\end{figure}

The experiments are performed in $1, 10, 50$ and $100$ epochs on FFT-based and non-FFT based algorithms. For the sake of space limit, we discuss the results for $100$ epochs only. The output obtained from the training is processed and clustered to identify the objects in the test image. For all the discussions on results, Figure \ref{fig:org} is used as one of the test images.

\begin{figure}[h!]
  \center
  \begin{subfigure}{0.5\linewidth}
      \center
\includegraphics[width=\linewidth]{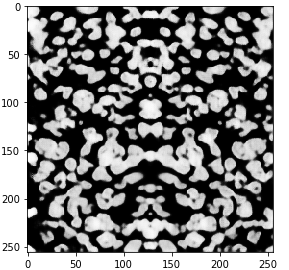}
\caption{With FFT.}
\label{fig:mskfft}
    \end{subfigure}%
  \begin{subfigure}{0.5\linewidth}
  \center
\includegraphics[width=\linewidth]{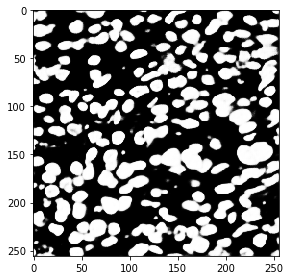}
\caption{without FFT.}
\label{fig:mskpfft}
    \end{subfigure}%
     \caption{Mask predicted.}
  \label{fig:mask}
\end{figure}

\begin{figure}[h!]
  \center
  \begin{subfigure}{0.5\linewidth}
      \center
\includegraphics[width=\linewidth]{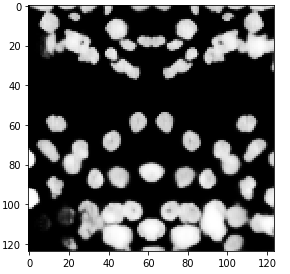}
\caption{with FFT.}
\label{fig:pfft}
    \end{subfigure}%
           \begin{subfigure}{0.5\linewidth}
      \center
\includegraphics[width=\linewidth]{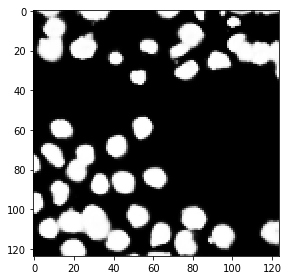}
\caption{Without FFT.}
\label{fig:wofft}
    \end{subfigure}%
     \caption{Unsampled prediction.}
  \label{fig:unsample}
\end{figure}

The images shown in Figures \ref{fig:mskfft} and  \ref{fig:mskpfft} are the output results when Figure  \ref{fig:org} is used as the test image. The mask prediction obtained using FFT-based algorithm is presented in Figure \ref{fig:mskfft} and Figure \ref{fig:mskpfft}, where the Figure \ref{fig:mskfft} is the predicted mask and Figure \ref{fig:pfft} is the unsampled prediction. The prediction obtained from model without FFT is shown in Figure \ref{fig:mskpfft} and unsampled prediction is Figure \ref{fig:wofft}.

Table \ref{tbl:ac_net} reports the performance metrics obtained for model in $100$ epochs with and without FFT, respectively. The evaluation metrics used in this experiment is Intersection Over Union (IoU) or the Jaccard index, a metric for measuring the accuracy of detecting objects. The average IoU obtained when FFT-based Neural Network is used is $0.83$; whereas, when FFT is not employed, the average IoU obtained is $0.52$. As the results suggest, the accuracy of object detection is much higher when FFT is used. 

\begin{table}[h]
\center
 \caption{Metric and cost function value for the experiment.}
\begin{tabular}{|l|c|c|c|c|c|}
\hline
\multicolumn{1}{|c|}{\bf FFT}	 & \multicolumn{1}{c|}{\bf Avg}	 & \multicolumn{1}{c|}{\bf DICE } &	\multicolumn{1}{c|}{\bf Mean} &	\multicolumn{1}{c|}{\bf CV. Val } &	\multicolumn{1}{c|}{\bf CV. Val } \\
  & \multicolumn{1}{c|}{\bf  Valid. IoU}	 & \multicolumn{1}{c|}{\bf Value} &	\multicolumn{1}{c|}{\bf  IoU} &	\multicolumn{1}{c|}{\bf  Loss} &	\multicolumn{1}{c|}{\bf  mean IoU} \\
\hline
With  &	0.83 &	-0.8392 &	0.8747	& -0.8753 &	0.8748 \\
    \hline
W/o & 0.52	& -0.5631	& 0.6815	& -0.6025 & 0.6815 \\
   \hline
\end{tabular}
\label{tbl:ac_net}
\end{table}

The loss function used in this experiment is DICE value \cite{dice1945measures} to assess the segmentation. 
The DICE value obtained with FFT based Algorithm is $-0.8392$ and without FFT based algorithm is $-0.5631$, showing a better similarity achieved by FFT. The Mean IoU for the final iteration obtained is $0.8747$ for FFT based algorithm and $0.6815$ for non-FFT based algorithm. 

For algorithm with FFT, cross-validation loss obtained is $-0.8753$ and for non-FFT based algorithm it is $-0.6025$. The cross-validation mean IoU is $0.8748$ and $0.6815$ for FFT based algorithm and non-FFT based algorithm, respectively.

\begin{figure}[h!]
  \center
  \begin{subfigure}{\linewidth}
      \center
\includegraphics[width=\linewidth]{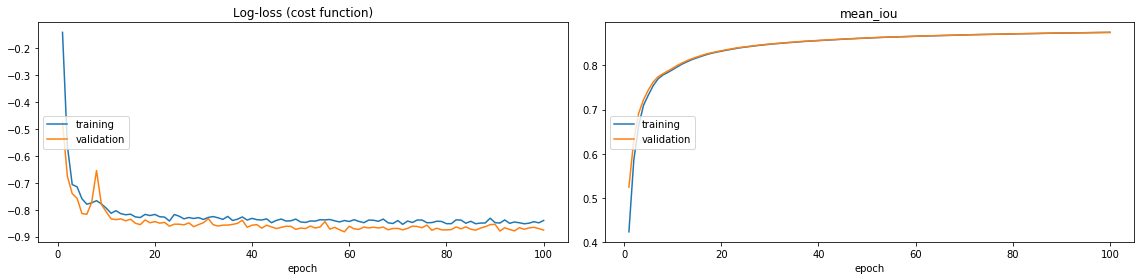}
\caption{With FFT.}
\label{fig:ll}
    \end{subfigure}%
    
           \begin{subfigure}{\linewidth}
      \center
\includegraphics[width=\linewidth]{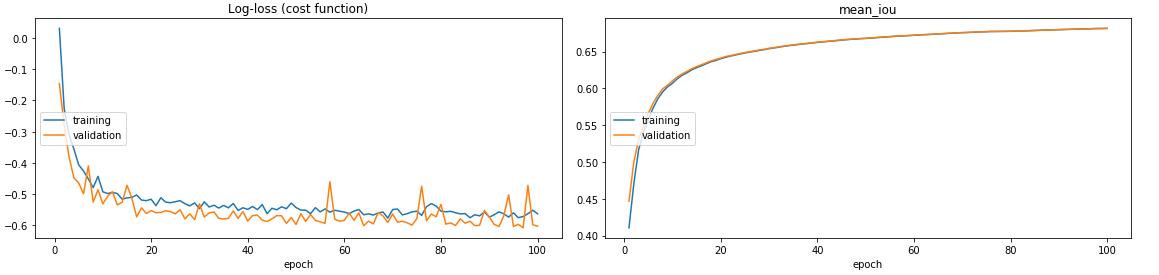}
\caption{without FFT.}
\label{fig:woll}
    \end{subfigure}%
     \caption{Log-Loss and mean IoU metric plots for 100 epochs.}
  \label{fig:logloss}
\end{figure}

Figures \ref{fig:ll} and \ref{fig:woll} illustrate the plots for loss and mean IoU obtained on the training and validation phases with FFT and without FFT, respectively. 
As log-loss plots illustrate the log-loss values are approaching to $-0.9$ and $-0.6$, respectively for with and without FFT, indicating the reduction in loss values. Moreover, it is also observable from the plots that the log-loss values are very stable and consistent when FFT-based approach is implemented; whereas, the log-loss values when FFT is not applied fluctuate between different epochs.

In the mean IoU metric plot, the X-axis consists the number of epochs and Y-axis consists of the mean  IoU value. The graph consists of plot for mean IoU value obtained for each epoch. Both FFT-based and non-FFT-based approaches show similar trends. However, as the scale of y-index indicates, the mean of IoU when FFT is utilized is much higher than compared to cases when FFT is not applied.

\subsection{Generate Local Maxima and Clustering Using DBScan}

\begin{figure}[h!]
\includegraphics[width=\linewidth]{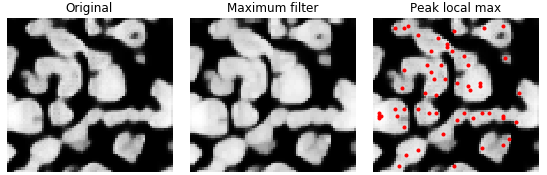}
\caption{Original image, Original image with maximum filter, Image with local peak maxima.}
\label{fig:mx}
\end{figure}

The input image for this experiment is obtained after the prediction of the test image from the model (Figure \ref{fig:mskfft}). The resultant image, Figure \ref{fig:mskfft}, is resized and used as the input. The local maxima are calculated based on the brightness of the pixel in the resized image, Figure \ref{fig:mx}, which will help in identifying peaks using the peak local maxima from scikit-image \cite{scikit-image}. 

In this experiment, Figure \ref{fig:mx} (left) is used as the input image to detect the local maxima by applying maximum filter to the original image where Figure \ref{fig:mx} (middle) is the image after maximum filter is applied. This helps in accentuating the high intensity region of the image. The local maxima function is applied to Figure \ref{fig:mx} (middle). The resultant image is Figure \ref{fig:mx} (right) where the high intensity pixels are identified using local maxima.

\begin{figure}[h!]
\center
\includegraphics[width=.6\linewidth]{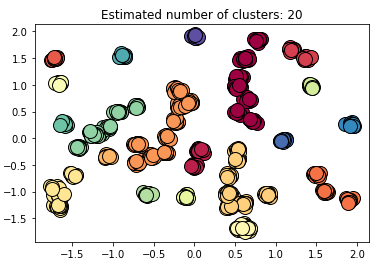}
\caption{DBScan cluster.}
\label{fig:dbs}
\end{figure}

The output image obtained with local maxima identified,  \ref{fig:mx} (right), are then clustered using DBScan to identify the number of objects by the number of clusters. 
Figure \ref{fig:dbs} shows the result of applying clustering using DBScan on Figure \ref{fig:mx}. The clustering algorithm has identified 20 clusters in the image.

\section{Conclusion and Future Work}
\label{sec:con}
In this work, an FFT-based convolutional neural network was successfully trained to identify the objects present in an image. The proposed FFT based convolutional neural network was able to improve the training time during convolution by reducing the number of steps needed to learn the feature and train the network. By comparing the Intersection Over Union (IoU) scores, we were able to determine the efficiency of the detection. 
The FFT was able to improve the training time during convolution from $600$-$700$ ms/step to $400$-$500$ ms/step. 
We compared FFT based model with non-FFT based model there and noticed an improvement in the Intersection Over Union (IoU) scores. In our work, for 100 epoch FFT based convolution IoU score was $0.83$ and for non-FFT based model the IoU score was $0.52$. 
The comparison of IoU scores implies that FFT based CNN has a better accuracy when compared to non-FFT CNN.

In conclusion we can say FFT-based convolution implemented in a CNN architecture can improve the convolution time and accuracy. This work can be done by comparing the model using multiple metrics and comparing the current model with many different object detection models.  It can also include comparing the effects of the model using multiple datasets with varied objects. This study can be further improved by finding the object position on a 3D environment to identify its position more precisely.

These results are promising and warrant further investigation. The DFFT can be implemented as a single layer NN \cite{velik_discrete_2008}, so it would seem that U-Net would be capable of learning this operation if it were useful for segmentation. Why U-Net is unable to transform or utilize the frequency domain of an image on its own is unclear but it is clear that performing the FFT at the onset is advantageous for the learning process.  

\section*{Acknowledgment}
This research work is supported by National Institute of Health (NIH) under Grants No: R15GM120669. The content is solely the responsibility of the authors and does not necessarily represent the official views of the National Institutes of Health. 

\bibliography{references}{}
\bibliographystyle{plain}
\end{document}